\documentclass[10pt,twocolumn,letterpaper]{article}

\usepackage{cvpr}
\usepackage{times}
\usepackage{epsfig}
\usepackage{graphicx}
\usepackage{amsmath}
\usepackage{amssymb}
\usepackage{algorithmic}
\usepackage{algorithm}
\usepackage{memhfixc}
\usepackage{amstext}
\usepackage{cite}
\usepackage{booktabs}	
\usepackage{fancyhdr}

\dblfloatsep\floatsep
\dbltextfloatsep\textfloatsep
\makeatletter
\renewcommand\normalsize{%
  \@setfontsize\normalsize\@xpt\@xiipt
  \abovedisplayskip 6\p@ \@plus2\p@ \@minus3\p@
  \abovedisplayshortskip \z@ \@plus3\p@
  \belowdisplayshortskip 4\p@ \@plus3\p@ \@minus3\p@
  \belowdisplayskip \abovedisplayskip
  \let\@listi\@listI}
\normalsize

\usepackage[pagebackref=true,breaklinks=true,letterpaper=true,colorlinks,bookmarks=false]{hyperref}

\cvprfinalcopy 

\ifcvprfinal\fi

\begin{document}


\title{Generalized Boundaries from Multiple Image Interpretations}

\author{
\begin{tabular}{ccc}
Marius Leordeanu$^{1}$ & Rahul Sukthankar$^{3,4}$ & Cristian Sminchisescu$^{2,1}$\\
{\small marius.leordeanu@imar.ro} & {\small rahuls@cs.cmu.edu} & {\small cristian.sminchisescu@ins.uni-bonn.de}\\
\end{tabular}\\
\\
$^1$Institute of Mathematics of the Romanian Academy\\
$^2$Faculty of Mathematics and Natural Science, University of Bonn\\
$^3$Google Research\\
$^4$Carnegie Mellon University\\
}

\maketitle

\thispagestyle{fancy}

\begin{abstract}

Boundary detection is essential for a variety
of computer vision tasks such as segmentation and recognition.
In this paper we propose a unified formulation and a novel
algorithm that are applicable to the detection of
different types of boundaries, such as intensity edges,
occlusion boundaries or object category specific boundaries.
Our formulation leads to a simple method with state-of-the-art
performance and significantly lower computational cost than existing methods.
We evaluate our algorithm on different types of boundaries, from low-level
boundaries extracted in natural images,
to occlusion boundaries obtained using motion cues and
RGB-D cameras, to boundaries from soft-segmentation.
We also propose a novel
method for figure/ground soft-segmentation that can be used
in conjunction with our boundary detection method and
improve its accuracy at almost no extra computational cost.

\end{abstract}

\section{Introduction}

Boundary detection is a fundamental problem in computer vision and
has been studied since the early days of the field.
The majority of papers on boundary detection
have focused on using only low-level cues, such as pixel intensity or
color~\cite{Roberts,Prewitt70,pb_marr,ref:Canny-86,pb_ruzon}.
Recent work has started exploring the problem
of boundary detection using higher-level representations of the image,
such as motion, surface
and depth cues~\cite{pb_Stein,pb_sundberg_cvpr11,pb_he_eccv10},
segmentation~\cite{pb_global}, as well as
category specific information~\cite{key:mairal_08,pb_malik_iccv11}.

\begin{figure}[ht]
\begin{center}
\includegraphics[scale = 0.48, angle = 0, viewport = 0 0 700 610, clip]{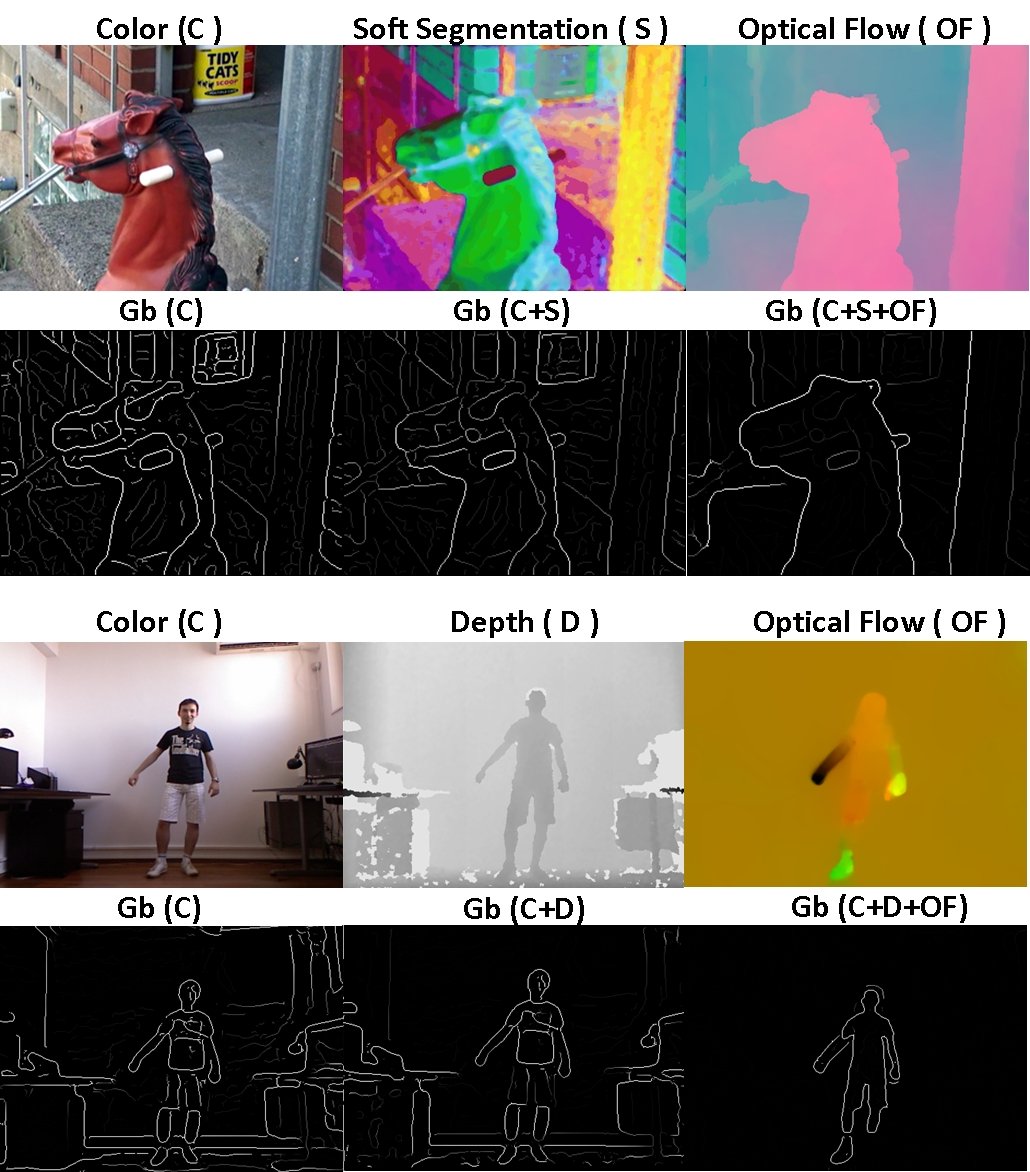}
\caption{Detection of occlusion and motion boundaries using the
proposed generalized boundary detection method (Gb).
First two rows: the input layers consist of
color~(C), soft-segmentation~(S) [the first three dimensions are shown
as RGB], and optical flow~(OF). Last two rows: input layers are
color~(C), depth~(D) and optical flow~(OF).  The same
implementation is used for both; combining multiple input layers
using Gb improves boundary detection. Best viewed in color.}
\label{fig:pb_diff_layers}
\end{center}
\end{figure}

In this paper we propose a general
formulation for boundary detection that
can be applied, in principle, to the identification of
any type of boundaries, such as general boundaries
from low-level static cues,
motion boundaries or category-specific boundaries
(Figures~\ref{fig:pb_diff_layers},
\ref{fig:results_examples},
\ref{fig:segPascal}).
Our method can be seen both as a generalization
of the early view of boundaries as step
edges~\cite{pb_review_koschan}, and as a unique closed-form solution to
current boundary detection problems, based on a straightforward mathematical
formulation.

We generalize the classical
view of boundaries from sudden signal changes on the original low-level image input~\cite{Roberts,Prewitt70,pb_marr,ref:Canny-86,pb_kanade,pb_di_zenzo,pb_cumani},
to a locally linear (planar or step-wise) model on multiple
layers of the input. The layers are interpretations of the image at different
levels of visual processing, which could be high-level (e.g., object category segmentation) or
low-level (e.g., color or grey level intensity).

Despite the abundance of
research on boundary detection, there is no general
formulation of this problem. In this paper, we
make the popular but implicit intuition of boundaries explicit:
boundary pixels mark the transition from one relatively constant
region to another, in appropriate interpretations of the image.
%
%
Thus, while the region constancy assumption may only apply weakly
for low-level input such as pixel intensity, it will also be weakly
observed in higher-level interpretation layers of the image.
Generalized boundary detection aims to exploit such weak signals
across multiple layers in a principled manner.
We could say that boundaries do not exist in the raw image,
but rather in the multiple interpretation layers of that image.
We can summarize our assumptions as follows:

\begin{enumerate}
\item  A boundary separates different image regions, which in the
    absence of noise are almost constant,
    at some level of image interpretation or processing.
    For example, at the lowest level, a region could
    have a constant intensity. At a higher-level, it could be a region delimitating an object category,
    in which case the output of a category-specific classifier would be constant.
\item For a given image, boundaries in one layer often coincide, in terms
of position and orientation, with boundaries in other layers. For example,
discontinuities in intensity are typically correlated with
discontinuities in optical flow, texture or other cues. Moreover,
the boundaries that align across multiple layers
often correspond to the semantic boundaries that are primarily of interest
to humans: the so-called ``ground-truth boundaries''.
\end{enumerate}

\noindent Based on these observations, we develop a unified model, which can
simultaneously consider both low-level and higher-level information.

Classical vector-valued techniques on
multi-images~\cite{pb_di_zenzo, pb_kanade, pb_review_koschan} can
be simultaneously applied to several
image channels, but differ from the proposed
approach in a fundamental way:
they are specifically designed for low-level input, by using first
or second-order derivatives of the image channels, with edge models limited to
very small neighborhoods of only a few pixels (for approximating the derivatives).
We argue that
in order to correctly incorporate higher-level information, one must go beyond a few pixels,
to much larger neighborhoods, in line with more recent methods~\cite{pb_global,pb_orig,pb_ren,pb_ruzon}.
First, even though boundaries from one layer coincide with edges from a
different layer, they cannot be required to match perfectly in location. Second,
boundaries, especially in higher-level layers, do not have to correspond to sudden
changes. They could be smooth transitions over larger regions and
exhibit significant noise that
would corrupt any local gradient computation.
That is why we advocate a linear boundary model rather than one based on
noisy estimation of derivatives, as discussed in the next
section.

Another drawback of traditional multi-image techniques is the issue
of channel scaling, where the algorithms require considerable
manual tuning.  Consistent with current machine learning based
approaches~\cite{pb_orig,pb_global,pb_dollar}, the parameters in our
proposed method are automatically learned using real-world datasets.
However, our method has better computational complexity and employs
far fewer parameters. This allows us to learn efficiently from limited
quantities of data without overfitting.

Another important advantage of
our approach over current methods
is in the closed-form computation of the boundary orientation.
The idea behind $Pb$~\cite{pb_orig} is to classify each possible boundary pixel based
on the histogram difference in color and texture information
between the two half disks on either side of a potential
orientation, for a fixed number of candidate angles (e.g., 8).
The separate computation for each orientation
significantly increases the computational cost and limits
orientation estimates to a particular granularity.



We summarize our contributions as follows:
1) we present a closed-form formulation of
generalized boundary detection that is computationally efficient;
2) we recover exact boundary normals through
direct estimation rather than evaluating coarsely sampled
orientation candidates;
3) as opposed to current approaches~\cite{pb_global,pb_sundberg_cvpr11},
our unified framework treats both low-level pixel data and
higher-level interpretations equally and can easily
incorporate outputs from new image interpretation algorithms;
and
4) our method requires learning only a single parameter per layer,
which enables efficient training with limited data.
We demonstrate the strength of our method on a variety of real-world tasks.


\section{Problem Formulation}

For a given $N_x \times N_y$ image $I$, let the $k$-th layer
$L_k$ be some real-valued array, of the same size, associated with $I$,
whose boundaries are relevant to our task.
For example, $L_k$ could contain, at each pixel,
the real-valued output of a patch-based binary classifier trained to detect man-made
structures or respond to a particular texture or color
distribution.\footnote{
The output of a discrete-valued multi-class classifier can be encoded as
multiple input layers, with each layer representing a given label.
}
Thus, $L_k$ will consist of relatively constant regions (modulo
classifier error) separated by boundaries. Note that the raw pixels
in the corresponding regions of the original image may not be constant.

Unlike some previous approaches, we expect that boundaries in
different layers may not precisely align.
Given a set of layers, each corresponding to a particular
interpretation level of the image, we wish to
identify the most consistent boundaries across multiple layers.
The output of our method for each point $\mathbf{p}$ on the
$N_x \times N_y$ image grid is a real-valued probability that
$\mathbf{p}$ lies on a boundary, given the information in all
multiple image interpretations $L_k$ centered at $\mathbf{p}$.

We model a boundary point in layer $L_k$ as
a transition (either sudden or gradual)
in the corresponding values of $L_k$ along the normal to the
boundary.
If several $K$ such layers are available,
let $\mathbf{L}$ be a three-dimensional array of size
$N_x \times N_y \times K$, such that $L(x,y,k) = L_k(x,y)$, for each $k$.
Thus, $\mathbf{L}$ contains all the relevant information
for the current boundary detection problem, given multiple
interpretations of the image or video.
Figure~\ref{fig:pb_diff_layers} illustrates how
we improve the accuracy of boundary detection by combining
different useful layers of information,
such as color, soft-segmentation and optical flow,
in a single representation $\mathbf{L}$,

Let $\mathbf{p}_0$ be the center of a window $W(\mathbf{p}_0)$ of size
$\sqrt{N_W} \times \sqrt{N_W}$. For each
image-location $\mathbf{p}_0$ we want to
evaluate the probability of boundary
using the information from $\mathbf{L}$, limited to that particular window.
For any $\mathbf{p}$ within the window, we make the following approximation,
which gives our locally linear boundary model:
\begin{equation}
\label{eq:boundary_approx}
L_k(\mathbf{p}) \approx C_k(\mathbf{p_0}) + b_k(\mathbf{p_0}) \mathbf{(\hat{p}_\epsilon - p_0)}^T\mathbf{n(p_0)}.
\end{equation}

\noindent
Here $b_k$ is nonnegative and corresponds to the
boundary ``height'' for layer $k$ at location $\mathbf{p_0}$;
$\mathbf{\hat{p}_\epsilon}$ is the closest point to $\mathbf{p}$
(projection of $\mathbf{p}$) on the disk of radius $\epsilon$ centered at
$\mathbf{p_0}$; $\mathbf{n(p_0)}$ is the normal to the boundary
and $C_k(\mathbf{p_0})$ is a constant over the window $W(\mathbf{p}_0)$.
This constant is useful for constructing our model (see Figure~\ref{fig:1D_model}),
but its value is unimportant, since it cancels out, as shown below.
Note that if we set $C_k(\mathbf{p_0}) = L_k(\mathbf{p_0})$
and use a sufficiently large $\epsilon$ such that
$\mathbf{\hat{p}_\epsilon = p}$,
our model reduces to the first-order Taylor expansion of $L_k(\mathbf{p})$
around the current $\mathbf{p_0}$.
\begin{figure}[ht]
\begin{center}
\includegraphics[scale = 0.37, angle = 0, viewport = 0 0 700 270, clip]{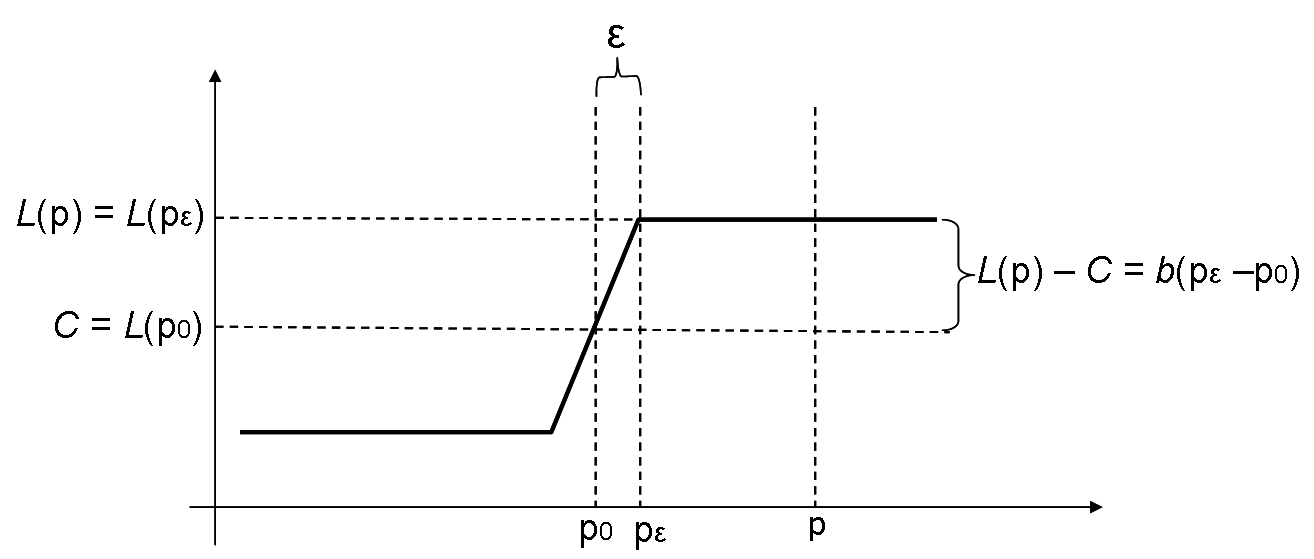}
\caption{Simplified $1$-dimensional view of our generalized boundary model. $\epsilon$ controls the region where the model is linear.
For points outside that region the layer is assumed to be roughly constant.}
\label{fig:1D_model}
\end{center}
\end{figure}

\begin{figure}[ht]
\begin{center}
\includegraphics[scale = 0.36, angle = 0, viewport = 0 0 700 250, clip]{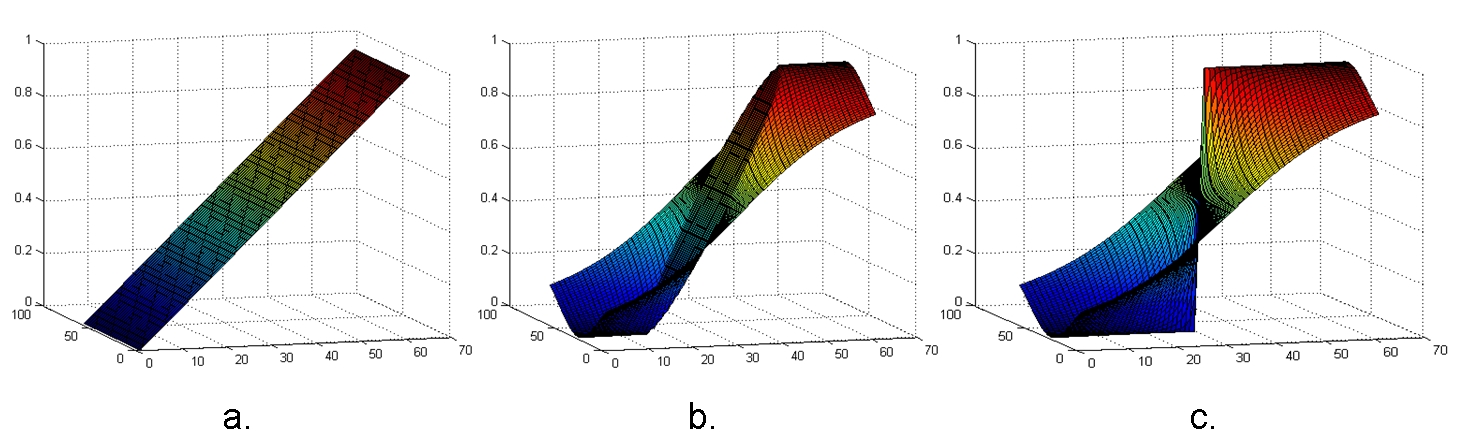}
\caption{Our boundary model for different values of $\epsilon$ relative to the window size $W$: a) $\epsilon > W$;
b) $\epsilon = W/2$ ; c) $\epsilon = W/1000$. When $\epsilon$ approaches zero the boundary model
becomes a step (along the normal direction passing through the window center).}
\label{fig:models_surf}
\end{center}
\end{figure}

As shown in Figures~\ref{fig:1D_model}
and~\ref{fig:models_surf}, $\epsilon$ controls the
steepness of the boundary, going
from completely planar when $\epsilon$ is large
(first-order Taylor expansion)
to a sharp step-wise discontinuity
through the window center $\mathbf{p_0}$, as $\epsilon$ approaches zero.
More precisely, when $\epsilon$ is very small
we have a step along the normal through
the window center, and a sigmoid which flattens as we get farther from the center, along the
boundary normal. As $\epsilon$ increases, the model flattens to become
a perfect plane for any $\epsilon$ that is larger than the window radius.

When the window is far from any boundary, the value of $b_k$ will be near zero,
since the only variation in the layer values is due to noise.
If we are close to a boundary, then
$b_k$ will become positive and large. The term $\mathbf{(\hat{p}_\epsilon - p_0)}^T\mathbf{n(p_0)}$ approximates the sign which indicates the side of
the boundary: it does
not matter on which side we are,
as long as a sign change occurs when the boundary
is crossed.


When a true boundary is present
within several layers at the same position~---
i.e., $b_k(\mathbf{p_0})$ is non-zero
and possibly different, for several $k$~---
the normal to
the boundary should be consistent.
Thus, we model the boundary normal $\mathbf{n}$
as common across all layers.

We can now write the above equation in matrix form for all layers, with the same
window size and location as follows.
Let $\mathbf{X}$ be a $N_W \times K$ matrix with a row $i$
for each location $\mathbf{p_i}$ of the window and a column for each layer $k$,
such that $X_{i;k}= L_k(\mathbf{p}_i)$.
Similarly, we define $N_W \times 2$ position matrix $\mathbf{P}$: on its $i$-th row we store the
$x$ and $y$ components of $\mathbf{(\hat{p}_\epsilon - p_0)}$ for
the $i$-th point of the window. Let $\mathbf{n}= [n_x, n_y]$ denote
the boundary normal and $\mathbf{b} = [b_1, b_2,\ldots , b_K]$ the step sizes
for layers $1, 2, \ldots, K$. Also, let us define
the rank-$1$ $2 \times K$ matrix $\mathbf{J = n}^T\mathbf{b}$. We also define matrix
$\mathbf{C}$ of the same size as $\mathbf{X}$, with each column $k$ constant and
equal to $C_k(\mathbf{p_0})$.

We can then rewrite Equation~\ref{eq:boundary_approx} as follows
(dropping the dependency on $\mathbf{p_0}$ for notational simplicity), with
unknowns $\mathbf{J}$ and $\mathbf{C}$:
\begin{equation}
\mathbf{X \approx C  + PJ} .
\end{equation}

Since $\mathbf{C}$ is a matrix with constant columns,
and each column of $\mathbf{P}$ sums to $0$, we have
$\mathbf{P}^T\mathbf{C}=\mathbf{0}$.
Thus, by multiplying both sides of the equation above
by $\mathbf{P}^T$ we can eliminate the unknown
$\mathbf{C}$. Moreover, it can be easily shown
that $\mathbf{P}^T\mathbf{P} = \alpha \mathbf{I}$,
i.e., the identity matrix scaled by a factor $\alpha$,
which can be computed since $\mathbf{P}$ is known. We
finally obtain a simple expression for the unknown
$\mathbf{J}$ (since both $\mathbf{P}$ and $\mathbf{X}$ are known):

\begin{equation}
\label{final_eq_1}
\mathbf{J} \approx \frac{1}{\alpha}\mathbf{P}^T\mathbf{X}.
\end{equation}

Since $\mathbf{J = n}^T\mathbf{b}$ it follows that $\mathbf{JJ}^T = \mathbf{\|b\|^2n}^T\mathbf{n}$
is symmetric and has rank $1$. Then $\mathbf{n}$ can be estimated as
the principal eigenvector of $\mathbf{M = JJ}^T$ and $\|\mathbf{b}\|^2$ as its largest eigenvalue.
$\|\mathbf{b}\|$, which is obtained as the square root of the largest eigenvalue of $\mathbf{M}$,
is the norm of the boundary steps vector $\mathbf{b} = [b_1, b_2, ..., b_K]$. This norm captures the
overall strength of boundaries from all layers simultaneously. If layers are properly scaled,
then $\|\mathbf{b}\|$ could be used as a measure of boundary strength.
Besides the intuitive meaning of $\|\mathbf{b}\|$, the spectral approach to boundary estimation
is also related to the gradient of multi-images previously
used for low-level color edge detection from classical papers such as~\cite{pb_kanade,pb_di_zenzo}.
However, it is important to notice that unlike those methods, we do
not compute derivatives, as
they are not appropriate for
higher-level layers and can be noisy for low-level layers.
Instead, we fit a model,
which, by controlling
$\epsilon$, can vary from planar to sigmoid/step-wise.
For smoother-looking results, in practice we weigh the rows of matrices $\mathbf{X}$ and $\mathbf{P}$ by a 2D Gaussian
with the mean set to the window center $\mathbf{p_0}$ and the standard deviation
equal to half of the window radius.

Once we identify $\|\mathbf{b}\|$, we pass it through a one-dimensional logistic model to obtain
the probability of boundary, similarly to recent classification approaches to boundary detection~\cite{pb_orig,pb_global}.
The parameters of the logistic regression model are learned using standard procedures.
The normal to the boundary $\mathbf{n}$ is
then used for non-maxima suppression.


\section{Algorithm and Numerical Considerations}

Before applying the main algorithm
we scale each layer in
$\mathbf{L}$ according to its importance, which may be problem dependent.
For example, in Figure~\ref{fig:pb_diff_layers}, it is clear that
when recovering occlusion boundaries, the optical flow layer (OF)
should contribute more than the raw color (C) and color-based soft
segmentation (S) layers. The images displayed
are from the dataset of Stein and Hebert~\cite{pb_Stein}.
The optical flow shown is an average between the flow~\cite{of_sun}
computed over two pairs of images:
(reference frame, first frame), and (reference frame, last frame).
We learn the correct scaling of the layers from training data
using a standard unconstrained nonlinear optimization procedure
(e.g., \emph{fminsearch} routine in MATLAB)
on the average $F$-measure of the training set.
We apply the same learning procedure in all of our experiments.
This is computationally feasible since there is
only one parameter per layer in the proposed model.

\begin{algorithm}
\caption{Gb1: Fast Generalized Boundary Detection}
\label{alg:bd_method}
\begin{algorithmic}
\STATE Initialize $\mathbf{L}$, scaled appropriately.
\STATE Initialize $w_0$ and $w_1$.
\FORALL {pixels $\mathbf{p}$}
  \STATE $\mathbf{M \gets (P}^T\mathbf{X_p)(P}^T\mathbf{X_p)}^T$
  \STATE $(\mathbf{v},\lambda) \gets $ principal eigenpair of $\mathbf{M}$
  \STATE $b_\mathbf{p} \gets \frac{1}{1+\exp(w_0+w_1\sqrt{\lambda})}$
  \STATE $\theta_\mathbf{p} \gets \text{atan2}(v_y, v_x)$
\ENDFOR
\RETURN {$\mathbf{b}$, $\mathbf{\theta}$}
\end{algorithmic}
\end{algorithm}

Algorithm~\ref{alg:bd_method} (referred to as Gb1)
summarizes the proposed approach.
The overall complexity of our method is relatively
straightforward to compute.
For each pixel $\mathbf{p}$,
the most expensive step
is the computation of the matrix $\mathbf{M}$, which takes
$O((N_W+2)K)$ steps
($N_W$ is the number of pixels in the window, and $K$ is the number of layers).
Since $\mathbf{M}$ is always $2 \times 2$, computing its eigenpair
$(\mathbf{v},\lambda)$ is a closed-form operation, with
a small fixed cost. It follows that
for a fixed window size $N_W$
and a total of $N$ pixels per image the overall complexity of our algorithm is $O(KN_WN)$. If
$N_W$ is a constant fraction $f$ of $N$, then complexity becomes $O(fKN^2)$.

Thus, the running time of Gb1 compares very favorably to that of
the $Pb$ algorithm~\cite{pb_orig,pb_global}, which in its exact form
has complexity $O(fKN_oN^2)$, where $N_o$ is a discrete number of
candidate orientations.
An approximation is proposed in~\cite{pb_global} with $O(fKN_oN_bN)$ complexity where $N_b$ is the number of
histogram bins for the different image channels.
However, $N_oN_b$ is large in practice and significantly affects
the overall running time.

\begin{figure}[ht]
\begin{center}
\includegraphics[scale = 0.6, angle = 0, viewport = -50 0 600 220, clip]{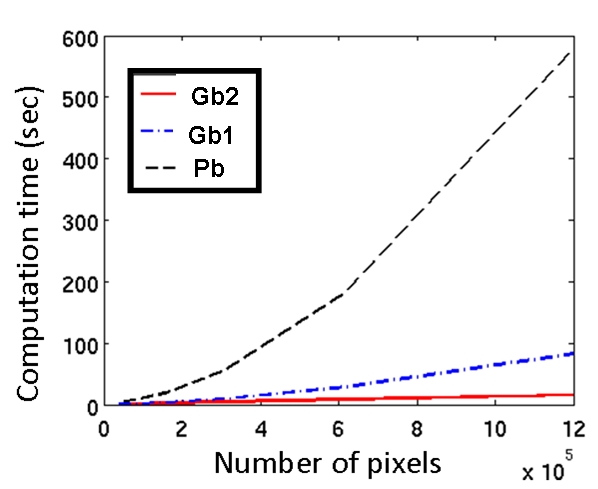}
\caption{ Edge detection
running times on a 3.2 GHz desktop of our non-optimized
MATLAB implementation of Gb1 and Gb2
vs.\ the publicly available code of
$Pb$~\cite{pb_orig}. Each algorithm uses the same window radius,
whose number of pixels is a constant fraction of the total
number of image pixels. Gb2 is linear in the number of image pixels
(independent of the window size). The accuracy of all algorithms is similar.
}
\label{fig:compTimes}
\end{center}
\end{figure}

We also propose a faster version of our algorithm, Gb2,
with complexity $O(fKN)$, that is \emph{linear} in the number of image pixels.
The speed-up is achieved by
computing $\mathbf{M}$ at a constant cost (independent of
the number of pixels in the window).
When $\epsilon$ is large and no Gaussian weighing is applied,
we have $\mathbf{P}^T\mathbf{X_p} = \mathbf{P_p^TX_p - P_0^TX_p}$, where $\mathbf{P_p}$ is the matrix of absolute
positions for each pixel $\mathbf{p}$ and $\mathbf{P_0}$ is a matrix with two constant columns equal to the
2D coordinates of the window center. Upon closer inspection, we note
that both $\mathbf{P}_\mathbf{p}^T\mathbf{X}$ and
$\mathbf{P_0}^T\mathbf{X}$ can be computed in constant time by using integral images, for each layer separately.
We implemented the faster version of our algorithm, Gb2, and verified experimentally that it is linear
in the number of pixels per image, independent of the window size (Figure~\ref{fig:compTimes}). The output
of Gb2 is similar to Gb1 (see Table~\ref{table:BSDS}),
and provably identical when $\epsilon$ is larger than the window radius
and no Gaussian weighting is applied. The weighting can be approximated by running Gb2 at multiple scales
and combining the results.

In Figure~\ref{fig:compTimes} we present
a comparison of the running times of edge detection
in MATLAB of the three algorithms (Gb1, Gb2 and
$Pb$~\cite{pb_orig}) vs.\ the number of pixels per image.\footnote{
Our optimized C++ implementation of Gb1 is an order of magnitude faster
than its MATLAB version.}

It is important to note that while our algorithm is fast,
obtaining some of the layers may be slow, depending on
the image processing required.
If we only use low-level interpretations, such as raw color or depth
(e.g., from an RGB-D camera) then the total execution time is small,
even for a MATLAB implementation.  In the next section, we propose
an efficient method for color-based soft-segmentation of images that
works well with our algorithm.  More complex, higher-level
inputs, such as class-specific segmentations naturally
increase the total running time.


\begin{figure*}[ht]
\begin{center}
\includegraphics[scale = 0.75, angle = 0, viewport = 0 0 700 250, clip]{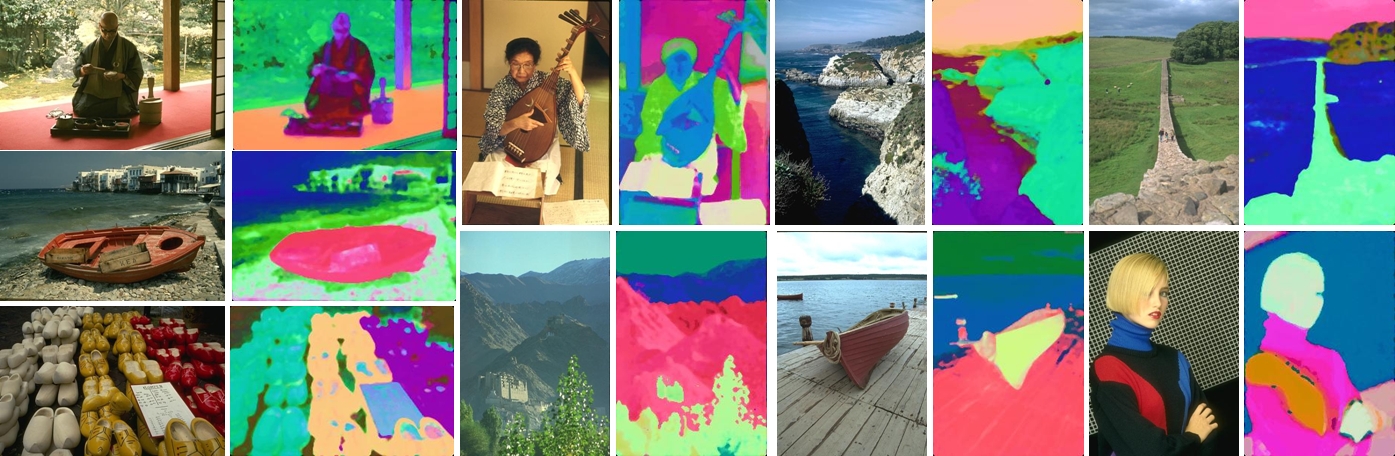}
\caption{Soft-segmentation examples using our method. The first three dimensions of the soft-segmentations, reconstructed using PCA,
are shown on the RGB channels. Total computation time for segmentation is less than $3$ seconds in MATLAB per image. Best viewed in color.}
\label{fig:soft_seg_examples}
\end{center}
\end{figure*}

\section{An Efficient Soft-Segmentation Method}
\label{sec:mySeg}

In this section we present a novel
method to rapidly generate
soft figure/ground image segmentations. Its soft continuous
output is similar to the eigenvectors computed by normalized cuts~\cite{ShiMa00}
or the soft figure/ground assignment obtained by alpha-matting~\cite{alpha_levin},
but it is much faster than most existing segmentation methods.
We describe it here because it
serves as a fast mid-level interpretation of the
image that significantly improves accuracy over raw color alone.

While we describe our approach in the context of color
information, the proposed method is general enough to handle a
variety of other types of low-level information as well.
The method is motivated by the observation that regions of
semantic interest (such as objects) can often be modeled with a
relatively uniform color distribution.  Specifically, we assume
that the colors of any image patch are generated from a
distribution that is a linear combination (or mixture) of a
finite number of color probability distributions belonging to the
regions of interest/objects in the image.

Let $\mathbf{c}$ be an indicator vector associated with some patch from the image,
such that $c_i = 1$ if color $i$ is present in the patch
and $0$ otherwise. If we assume that the image is formed by a composition of regions
of uniform color distributions, then we can consider $\mathbf{c}$ to be a multi-dimensional
random variable drawn from a mixture (linear combination)
of color distributions $\mathbf{h}_i$
corresponding to the image regions:

\begin{equation}
\mathbf{c} \sim \sum_i \pi_i \mathbf{h}_i .
\end{equation}

The linear subspace of color distributions can be automatically discovered by performing PCA on
collections of such indicator vectors $\mathbf{c}$, sampled uniformly from the image.
This idea deserves a further in-depth discussion but, due to space limitations,
in this paper we outline just
the main idea, without presenting our detailed probabilistic analysis.

Once the subspace is discovered using PCA,
for any patch sampled from the image and its associated indicator vector
$\mathbf{c}$, its generating
distribution (considered to be the distribution of the foreground)
can be reconstructed
from the linear subspace using the usual PCA reconstruction approximation:
$\mathbf{h_F(c) \approx h_0 + \sum_i (c-h_0)}^T\mathbf{v_i}$. The distribution of the
background is also obtained from the PCA model using the
same coefficients, but with opposite sign.
As expected, we obtain a background
distribution that is as far as possible (in the subspace) from
the distribution of the foreground: $\mathbf{h_B(c) \approx h_0 - \sum_i (c-h_0)}^T\mathbf{v_i}$.

Using the figure/ground distributions obtained in this manner,
we classify each point in the image as either belonging or not
to the same region as the current patch.
If we perform the same classification procedure
for $n_s$ ($\approx 150$) locations uniformly sampled on the image grid, we
obtain $n_s$ figure/ground segmentations for the same image. At a final step,
we again perform PCA on vectors collected from all pixels in the
image; each vector is of dimension $n_s$ and corresponds to a certain
image pixel, such that its $i$-th element is equal
to the value at that pixel in the $i$-th
figure/ground segmentation. Finally we perform PCA reconstruction using the first $8$
principal components, and obtain a set of $8$ soft-segmentations which are a compressed
version of the entire set of $n_s$ segmentations. These soft-segmentations
are used as input layers to our boundary detection method, and are similar
in spirit to the normalized cuts eigenvectors computed for $gPb$~\cite{pb_global}.

In Figure~\ref{fig:soft_seg_examples} we show examples of
the first three such soft-segmentations on the RGB color channels.
This method takes less than $3$ seconds in MATLAB on a
3.2GHz desktop computer for a $300 \times 200$ color image.

\section{Experimental analysis}
\label{sec:results}

To evaluate the generality of our proposed method, we conduct
experiments on detecting boundaries in image, video and RGB-D
data on both standard and new datasets.
First, we test our method on static color images for which we only
use the local color information.
Second, we perform experiments on occlusion boundary detection
in short video clips. Multiple frames, closely spaced in time,
provide significantly more information
about dynamic scenes and make occlusion boundary detection possible, as shown
in recent work~\cite{pb_Stein,pb_sundberg_cvpr11,pb_sargin,pb_he_eccv10}. Third, we also
experiment with RGB-D images of people and show that the depth layer can be effectively
used for detecting occlusions. In the fourth set of experiments we use the
CPMC method~\cite{CPMC_2010}
to generate figure/ground category segments on the PASCAL2011 dataset.
We show how it can be effectively used to generate image layers that can produce
high-quality boundaries when processed using our method.

\subsection{Boundaries in Static Color Images}

\begin{table}
\caption{Comparisons of accuracy (F-measure) and computational time between our method and two other
popular methods on BSDS dataset. We use two versions of the proposed method: Gb1 (S) uses color and soft-segmentations as
input layers, while Gb1 uses only color. Color layers are represented in CIE Lab space.}
\label{table:BSDS}
\begin{center}
\begin{tabular}{lccccc}
\toprule
 Algorithm       & Gb1 (S)                & Gb1           & Gb2 & Pb~\cite{pb_orig}  & Canny~\cite{ref:Canny-86}   \\
\midrule
  F-measure  & \textbf{0.67}   & 0.65           & 0.64     & 0.65       & 0.58                                \\
  Time (sec) & 8               & 3              & 2        & 20       & 0.1                                 \\
\bottomrule
\end{tabular}
\end{center}
\end{table}

We evaluate our proposed method on the well-known BSDS300
benchmark~\cite{pb_orig}. We compare the accuracy and computational time of
Gb with $Pb$~\cite{pb_orig} and Canny~\cite{ref:Canny-86}
edge detector. All algorithms use only local information at a single scale.
Canny uses brightness information, Gb uses brightness and color, while
$Pb$ uses brightness, color and texture information.
Table~\ref{table:BSDS} summarizes the results.
Note that our method is much faster than $Pb$ (times are averages in Matlab on the same
$3.2$ GHz desktop computer). When no texture information is used for $Pb$, its accuracy drops significantly
while the computational time remains high ($\approx 16$ seconds).
%

\subsection{Occlusion Boundaries in Video}

\begin{table}
\caption{Performance comparison on the CMU Motion Dataset of current
techniques for occlusion boundary detection.}
\label{table:CMU-Stein}
\begin{center}
\begin{tabular}{lr}
\toprule
   Algorithm                                  & F-measure    \\
\midrule
  Gb1                                       & \textbf{0.63} \\
  He et al.~\cite{pb_he_eccv10}              & 0.47          \\
  Sargin et al.~\cite{pb_sargin}             & 0.57          \\
  Stein et al.~\cite{pb_Stein}              & 0.48          \\
  Sundberg et al.~\cite{pb_sundberg_cvpr11}  & 0.62          \\
\bottomrule
\end{tabular}
\end{center}
\end{table}

Occlusion boundary detection is an important problem and has
received increasing attention in computer vision.
Current state-of-the-art techniques are based on the computation of optical flow combined with a
global processing phase~\cite{pb_Stein,pb_sundberg_cvpr11,pb_sargin,pb_he_eccv10}.
We evaluate our approach on the CMU Motion Dataset~\cite{pb_Stein}
and compare our method with published results on the same dataset
(summarized in Table~\ref{table:CMU-Stein}).
Optical flow is an important cue for detecting occlusions in video;
we use Sun et al.'s publicly available code~\cite{of_sun}.
In addition to optical flow, we provided Gb-1 with two additional
layers: color and our soft segmentation (Section~\ref{sec:mySeg}).
In contrast to the other
methods~\cite{pb_Stein,pb_sundberg_cvpr11,pb_sargin,pb_he_eccv10},
which require significant time for processing and optimization,
Gb requires less than $4$ seconds on average (aside from the
external optical flow routine) to process images ($230\times320$)
from the CMU dataset.


\subsection{Occlusion Boundaries in RGB-D Video}

\begin{figure*}[ht]
\begin{center}
\includegraphics[scale = 0.61, angle = 0, viewport = 0 0 800 450, clip]{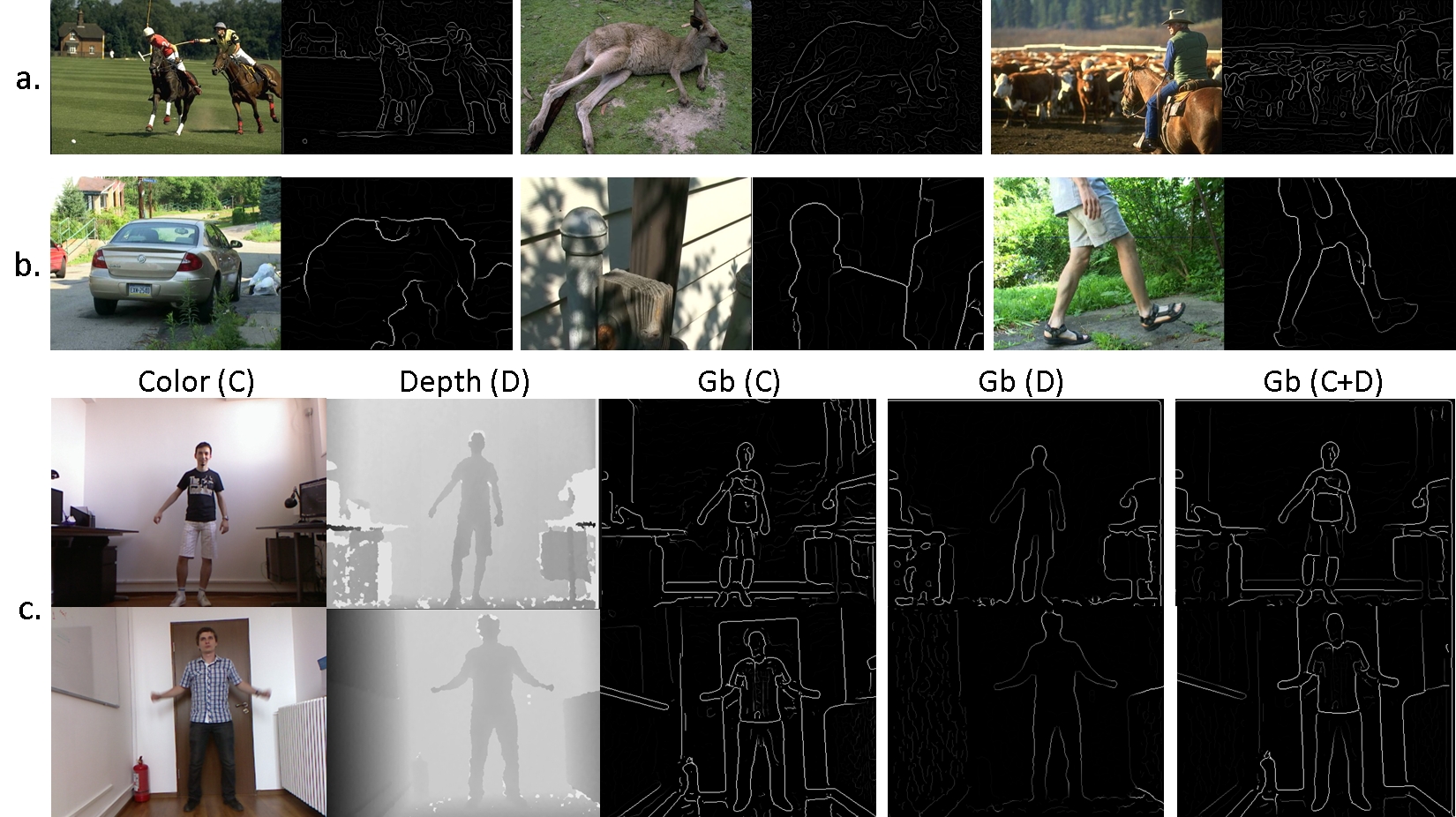}
\caption{Example results of Gb1 using different input layers: a) color and soft-segmentation
on BSDS300; b) color, soft-segmentation and optical flow on CMU Motion Dataset; c) color and depth from
RGB-D images.}
\label{fig:results_examples}
\end{center}
\end{figure*}

The third set of experiments uses RGB-D video clips of people performing different
actions. We combine the low-level color and depth input with large-displacement optical flow~\cite{ldof_brox09}, which is useful for large inter-frame body movements.
Figure~\ref{fig:pb_diff_layers} shows an example of the input layers and the output of our method.
The depth layer was pre-processed to retain the largest connected
component of pixels at a similar depth, so as to cover the main
subject performing actions.
Table~\ref{table:RGB-D} summarizes boundary detection in RGB-D on
our dataset of $74$ training and $100$ testing images.\footnote{
 We will release this dataset to enable direct comparisons.
}
We see that Gb can effectively combine information from
color~(C), optical flow~(OF) and depth~(D) layers to achieve better
results.
Figure~\ref{fig:results_examples}) shows sample
qualitative results for Gb using only the basic color and depth
information (without pre-processing of the depth layer).
Without optical flow, the total computation time for boundary
detection is less than $4$ seconds per image in MATLAB.

\begin{table}
\caption{Average F-measure on $100$ test RGB-D frames of Gb1 algorithm,
using different layers: color (C), depth (D) and
optical flow (OF). The performance improves as more layers are combined.
Note: the reported time for C+OF and C+D+OF does not include that of
generating optical flow using an external module.}
\label{table:RGB-D}
\begin{center}
\begin{tabular}{lccc}
\toprule
 Layers    & C+OF       & C+D      & C+D+OF\\
\midrule
  F-measure  & 0.41     & 0.58       &\textbf{0.61}  \\
  Time (sec) & 5        & 4          & 6             \\
\bottomrule
\end{tabular}
\end{center}
\end{table}



\subsection{Boundaries from soft-segmentations}

Our previous experiments use our soft-segmentation method
as one of the input layers for Gb. In all of our experiments, we find
the mid-level layer information provided by soft-segmentations significantly
improves the accuracy of Gb.

\begin{figure}
\begin{center}
\includegraphics[scale = 0.52, angle = 0, viewport = 0 0 500 570, clip]{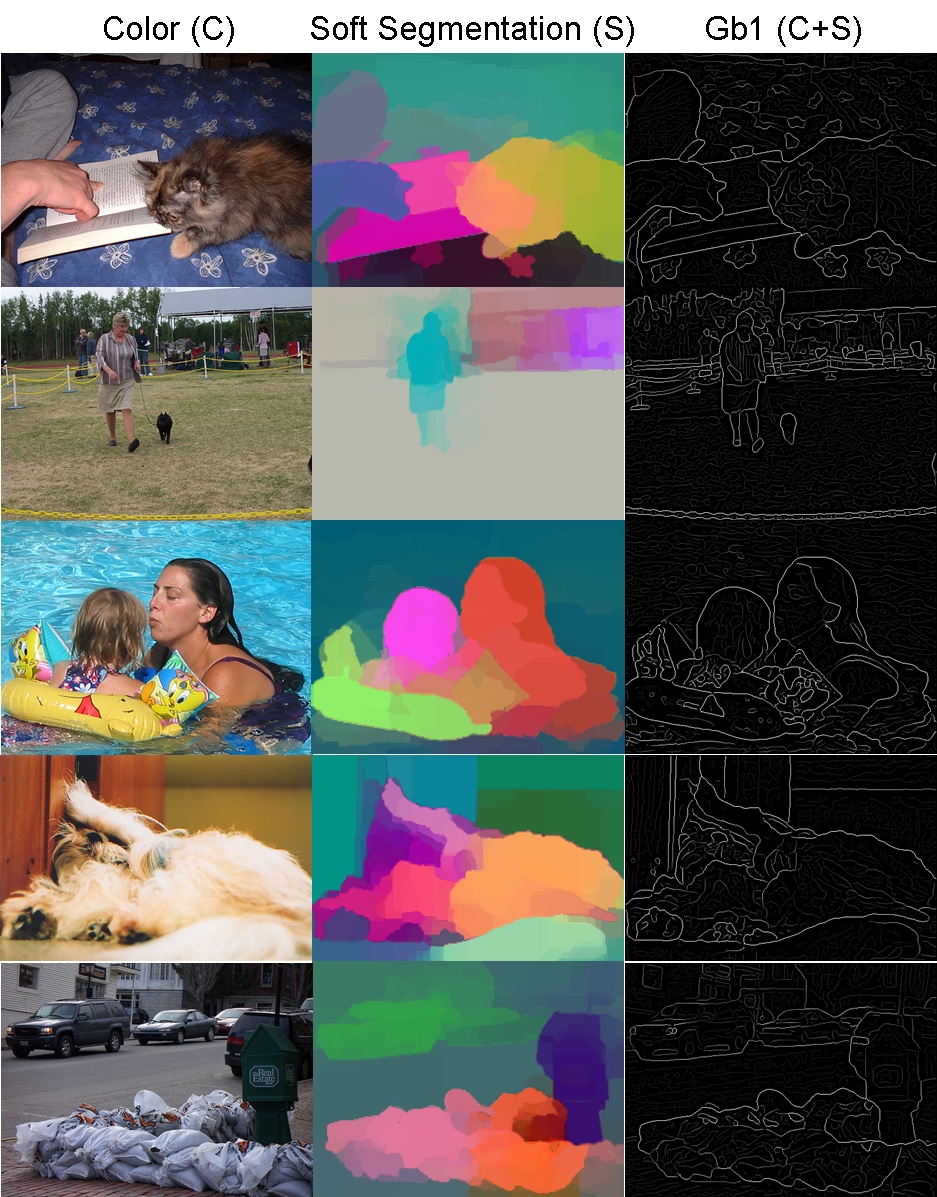}
\caption{Qualitative results using Gb on PASCAL2011 images, from color and soft-segmentations obtained
from the output of CPMC~\cite{CPMC_2010}. Best viewed on the screen.}
\label{fig:segPascal}
\end{center}
\end{figure}

The PCA reconstruction procedure described in Section~\ref{sec:mySeg}
can also be applied to a large pool of figure/ground segments, such as
those generated by the CPMC method~\cite{CPMC_2010}.
This enables us to achieve an F-measure of $0.70$ on
BSDS300, which matches the performance of $gPb$~\cite{pb_global}.
CPMC+Gb also gives very promising results on the PASCAL2011
dataset, as evidenced by the examples in Figure~\ref{fig:segPascal}.
These preliminary results indicate that fusing evidence from
color and soft-segmentation using Gb is a promising avenue for
further research.


\section{Conclusions}

We present Gb, a novel model and algorithm for generalized boundary detection.
Our method effectively combines multiple
low- and high-level interpretation layers of an input image in a
principled manner to achieve state-of-the-art accuracy on
standard datasets at a significantly lower computational cost
than competing methods.
Gb's broad real-world applicability is demonstrated through
qualitative and quantitative results on detecting
semantic boundaries in natural images, occlusion boundaries in video
and object boundaries in RGB-D data.
We also propose a second, even more efficient variant of Gb,
with asymptotic computational complexity that is linear with
image size.  Additionally, we introduce a practical method for
fast generation of soft-segmentations, using
either PCA dimensionality reduction on data collected from
image patches or a large pool of figure/ground segments.
We also demonstrate experimentally that our soft-segmentations are
valuable mid-level interpretations for boundary detection.



{\small
\bibliographystyle{ieee}
\bibliography{complete}
}
\end{document}